\documentclass[acmtog, authorversion]{acmart}

\usepackage{booktabs} % For formal tables
\usepackage{textgreek}
\usepackage[ruled]{algorithm2e} % For algorithms
\usepackage{graphicx}
\usepackage{caption}
\usepackage{amsmath}
\usepackage{array,booktabs}
\usepackage{cite}

\begin{document}
% Title portion
\title{Quantification of Damage Using Indirect Structural Health Monitoring}
 \subtitle{Department of Civil and Environmental Engineering, Carnegie Mellon University}

\author{Achyuth Madabhushi}
\email{amadabhu@alumni.cmu.edu}

\begin{abstract}

Structural health monitoring is important to make sure bridges do not fail. Since direct monitoring can be complicated and expensive, indirect methods have been a focus on research. Indirect monitoring can be much cheaper and easier to conduct, however there are challenges with getting accurate results. This work focuses on damage quantification by using accelerometers. Tests were conducted on a model bridge and car with four accelerometers attached to to the vehicle. Different weights were placed on the bridge to simulate different levels of damage, and 31 tests were run for 20 different damage levels. The acceleration data collected was normalized and a Fast-Fourier Transform (FFT) was performed on that data. Both the normalized acceleration data and the normalized FFT data were inputted into a Non-Linear Principal Component Analysis (separately) and three principal components were extracted for each data set. Support Vector Regression (SVR) and Gaussian Process Regression (GPR) were used as the supervised machine learning methods to develop models. Multiple models were created so that the best one could be selected, and the models were compared by looking at their Mean Squared Errors (MSE). This methodology should be applied in the field to measure how effective it can be in real world applications.
\end{abstract}

\maketitle

\section{Introduction}

%Introduction:  formal  problem  statement  that  describe  the  background, motivation,  system  of  interest,  phenomena  of  interest,  operational  and environmental conditions, etc. EMPHASIZE your contributions and innovation.
\flushleft{
The objective of this project is to quantify bridge damage by employing indirect monitoring. This experiment was conducted on a laboratory model using four accelerometer sensors and a data acquisition system (DAQ). Information was collected in LABView and data processing was done using MatLab. Most current damage quantification methods use Principal Component Analysis (PCA) which does not account for non-linear relationships in the input data set. A Non-Linear PCA (NLPCA) was used to reduce the dimensionality, since in datasets with high dimensionality, statistical inference becomes difficult to achieve with linear methods.
\flushleft{
The testing data used to test the performance of the model would be 20\% of the initially collected data, through k-fold cross validation. Learning tools used in this analysis were Support Vector Regression (SVR) and Gaussian Process Regression (GPR). The goal is to achieve an accuracy within 10\% of the highest damage scenario (190 grams).}

\section{Literature Review}
\flushleft{Due to the cost and severity of bridge failures, the health monitoring of bridges has always been a topic of research. Much of the literature found has been focused on different ways to inspect bridges. Researchers have used different strain sensors and accelerometers to try and model bridge damage. Optical sensors such as the FBG sensor offer advantages over previous sensors such as their durability, compact size, and high precision [14]. Visual inspection remains one of the most common bridge inspection methods, however it is not an optimal method as it is subjective and can only be used to detect surface issues on areas that are visible [15]. A wide variety of highly effective local non-destructive evaluation tools are available for health monitoring like condition monitoring and damage prognosis [2].Charles's article focuses on the introduction and development of indirect SHM, including the advantages over other monitoring methods. To enrich the measures of SHM, Y.B. Yang's article on measuring bridge faults with vehicles was by deriving the dynamic equations and method of modal superposition to derive the solutions for the vehicle-bridge system, which consists of bridge response, vehicle response, and displacement responses.}
\flushleft{
Beginning in 2005, Y.B. Yang has produced a significant body of theoretical literature highlighting the relationships between acceleration and bridge damage. This provides the theoretical framework for the approach taken in this project to vehicle bridge interactions. The acceleration of a bridge is mostly dependent on the natural frequency component of vibration frequency [1]. This article is limited in the sense that the natural frequency changes of the bridge is mentioned as a potential method to monitor the speed of cars. On the contrary, this project plans to measure the changes in a vehicle's acceleration as a result of varying damage states of a bridge.}
\flushleft{
Lederman et.al.[16] performed experiments similar to this project in an effort to quantify damage. The main difference being that Lederman et.al used PCA as their feature extraction method. PCA is limited because it cannot extract non-linear relationships from the acceleration data.
}

\section{Description of Sensing System}

%Description  of  sensing  system  detailing  sensing,  signal  conditioning  and communication hardware requirements and issues as they relate to the problem. 

The experiment involved using a laboratory model as a bridge. The model was made of a vehicle with constant speed passing over a lab-scale bridge. Four uniaxial accelerometers were attached to the model car (two sensors on the chassis and two sensors on the wheel axles). The accelerometers used were Mistras Model 5102, which have a high sensitivity [3]. The bridge damage was modeled by varying weights at the same location on the bridge. As the model vehicle passed over the bridge, the accelerometers collected data and transmitted it to the computer via the DAQ. 31 runs were tested for each of the 20 different weights, resulting in 620 total runs. For each weight group, the model vehicle passed over the bridge forwards and backwards (so that it would go back to its starting position) 31 times, the raw data was collected throughout. Thus, there were 20 sets of raw data collected.

\begin{figure}[h!]
  \includegraphics[width=0.3\textwidth]{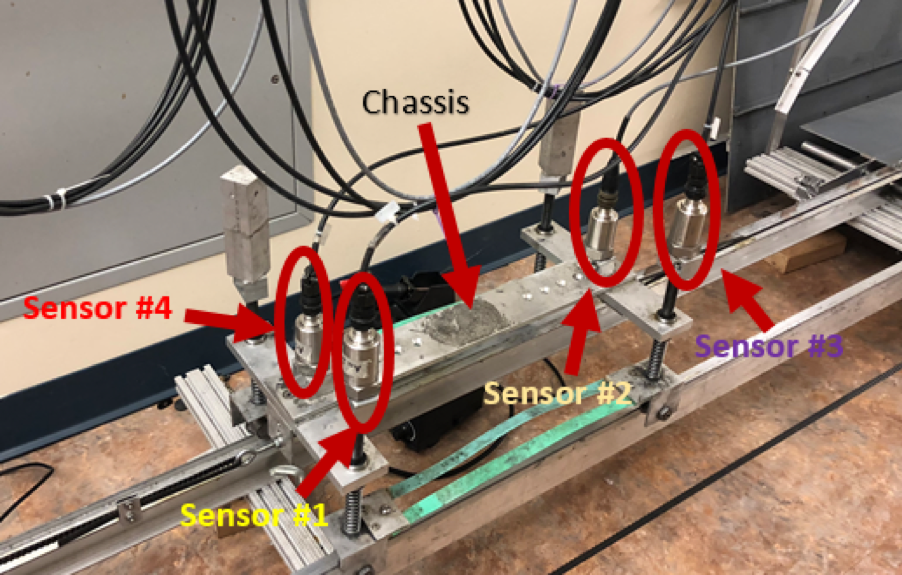}
  \caption{Model Vehicle Used in Testing}
\end{figure} 

\begin{figure}[h!]
  \includegraphics[width=0.3\textwidth]{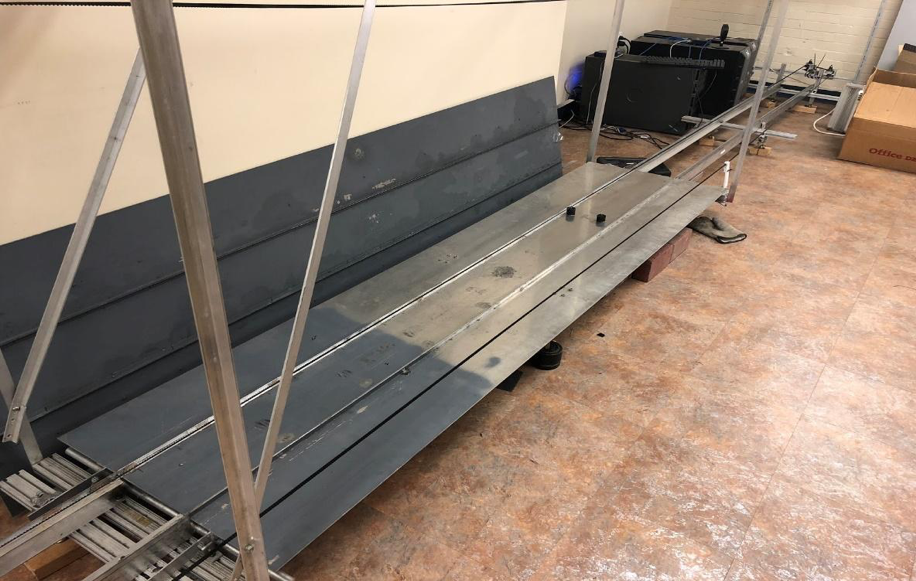}
  \caption{Model Bridge}
\end{figure} 

\section{Feature Extraction}
%Description of features, identifying the relevant features or information extracted from the measurements, comparing and contrasting their utility in terms of effectiveness, implementation, cost, computational effort, etc.This part should also discuss the ability (or limitations) of the feature(s) to accurately track the information in the presence of operational and environmental variability, and how you quantify when changes in the features are indicative of the change of interest. How is data normalization accomplished?  
\flushleft{

}
\flushleft{Current quantification efforts use PCA for linear feature extraction and as such reduce the accuracy of the data analysis. A non-linear feature extraction technique is used as opposed to PCA to improve the accuracy in the quantification of damage. Non-linear feature extraction allows features typically ignored by linear extraction methods to be collected.}

\flushleft{Non-linear PCA (NLPCA) reduces the dimensions, thus avoiding the curse of dimensionality. By reducing the dimensions, the complexity and computational time of the classification or regression algorithm can be reduced, as the input to the learning algorithm is simplified. These simpler models reduce the variance when testing the data set [5]. NLPCA incorporates both nominal and ordinal variables [12]. Nominal variables are the variables that do not have a quantifiable value, and ordinal variables are the variables that have different categories[4].}

\flushleft{The extracted features have reduced dimensionality while preserving the quality of the original data. Similar to linear PCA, the goal is to determine the nature of the original data set. NLPCA describes curved or multidimensional correlations within the original data sub-space. These nonlinear components are weighted summaries of the original variables [12]. NLPCA also allows for the easy identification of noise in the original dataset.  Fewer features allows for a better understanding of the original dataset.
}

\flushleft{NLPCA is used to extract three principal components from three datasets: the normalized acceleration data, the normalized Fast Fourier Transform (FFT) data and normalized wavelet transform data. The acceleration data was obtained by slicing the raw data files to create a matrix with the 31 forward runs for each weight. The acceleration data was normalized for each weight group by subtracting the mean value of all runs in that weight group from the acceleration value. The FFT was conducted on the normalized acceleration data, and then it was normalized by subtracting the minimum value of each weight group and dividing the result by the range of values. The same procedure was done for the wavelet transform data. The idea behind finding the principal components for three different types of data was to see which dataset would give the best results after machine learning was applied.

\section{Machine Learning Methods}
%Discuss the statistical models that are employed to discriminate features from different states of the system. What are you trying to achieve through the statistical  modeling?  How  are  thresholds  established,  if  any?   Do you use supervised or unsupervised learning approaches? What types of models are employed (classification, regression, outlier detection)? How will you “train” the models? 

\subsection{Regression Versus Classification}

\flushleft{A supervised learning method was chosen because since the values of the output variable (the weights) for the data collected is known. Since the output variable is known, a supervised model can be used to increase accuracy of prediction. The next decision was choosing whether a classified learning or regression model was more useful. Given the continuous nature of possible damage on a bridge, a regression model would be more apt. A classification model which would separate outputs into low, medium and high damage scenarios was considered. While this has its use, it would not be as practical as a regression model given that a regression model would allow the model output to be compared to a bridge's design failure load. Moving further, having high-dimensional data would necessitate appropriate regression models. Both GPR and SVR were used with a 5-fold cross validation.}

\subsubsection{Support Vector Regression}
\flushleft{
The basic idea of SVR involves determining a function f(x) which has at most {\textepsilon}  deviation from the actual target points y for all the training data [8]. The {\textepsilon} generates a range around the model prediction outside which all predictions are rejected as errors. SVR works well because of its ability to obtain sparse solutions [10]. {\textepsilon} is also known as the generalized error bound. Unlike other regression models that minimize the training error, SVR minimizes this generalized error bound to achieve general model performance. The generalized error bound is a combination of the training error and a regularization term [10]. 

\begin{align}
0.5*\sum \alpha_{i}*\alpha_{j}*G(x_{j},x_{k})\label{SVR_eqn}
\end{align}
%probably need to explain this a little more, also include other constraint function

SVR also has the ability of employing kernel functions, which allows for non-linear datasets to be fitted by the regression [7]. The kernel trick replaces the dot products (that are needed in SVR for the Gram Matrix) by the kernel function specified, which creates a feature space in which the regression method can be applied [8]. The Gram Matrix is used in the objective function along with {\textalpha} Lagrange multipliers as is shown in equation \ref{SVR_eqn} above.} For each of the three datasets, the principal components were input into an SVR function using three different Kernel functions: Linear, Polynomial and Gaussian.}
 
\subsection{Gaussian Process Regression}

\flushleft{ GPR assumes that the observed and predicted variables follow a multivariate Gaussian distribution. The Kernel function is used to create the covariance matrices from the x-values (features) that in turn predict values of the y-values (weight/damage) according to the magnitude of the features. This prediction is done by calculating the mean of the distribution for each of the x-values (Equation \ref{ya}) and the uncertainty of that prediction is the variance of the distribution at that x-value (Equation \ref{varya}). It should be noted that this project has three-dimensional inputs which means that the x-values used in the covariance functions are vectors and not scalars (Ebden 2015)\\}

\begin{align}
\textbf{K}  = \begin{bmatrix}
k(x_{1},x_{1}) & k(x_{1},x_{2}) & \dots & k(x_1,x_n) \\
k(x_{2},x_{1}) & k(x_{1},x_{2}) & \dots & k(x_2,x_n) \\
\vdots & \vdots & \ddots & \vdots \\
k(x_{n},x_{1}) & k(x_{n},x_{2}) & \dots & k(x_n,x_n)\\
\label{covar_mat}
\end{bmatrix}
\end{align}
\begin{align}
\mathbf{K_a} = \begin{bmatrix}
k(x_{a},x_{1}) & k(x_{a},x_{2}) & \dots & k(x_{a},x_n) \\
\label{ka}
\end{bmatrix}
\end{align}
\begin{align}
\mathbf{K_{aa}} = \begin{bmatrix}
k(x_{a},x_{a}) \label{kaa}
\end{bmatrix}
\end{align}

\flushleft{If $y_a$ is the mean of the distribution, it can be found as follows:}
\begin{align}
y_{a} = \mathbf{K_a}*\mathbf{K^{-1}}*\mathbf{y} \label{ya}
\end{align}

\flushleft{ Where $y_a$ is a vector of observed output values (weights). The variance of the distribution can be found as follows:}
\begin{align}
var(y_{a}) = \mathbf{K_{aa}}-\mathbf{K_a}*\mathbf{K_a^{-1}}*\mathbf{K^{T}} \label{varya}
\end{align}

\section{Results}

\subsection{Feature Extraction}
  \flushleft{The results of the dimensionality reduction are displayed in Figures 3-5. Figure 3 shows the principal components of normalized acceleration data. Figure 4 shows the principal components of normalized FFT data. For visualization, both figures shows the feature spectrum extracted from data collected from the back wheel accelerometer of the vehicle.}
%include the wavelet graph and explain how it is bad
	\flushleft{Figure 5 represents the principal components of normalized wavelet data. When comparing Figure 5 to Figures 3 and 4, it can be seen that the grouping clusters are not as easy to identify. Since the clustering was not as clear, the SVR results for this dataset were not expected to be very helpful.
}

\begin{figure}[h!]
  \includegraphics[width=0.4\textwidth]{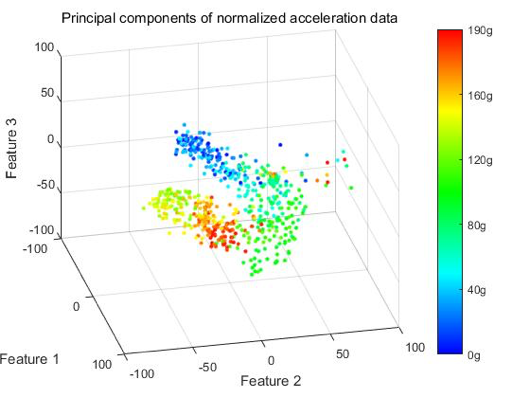}
  \caption{Principal Component of Normalized Acceleration Data }
\end{figure} 

\begin{figure}[h!]
  \includegraphics[width=0.4\textwidth]{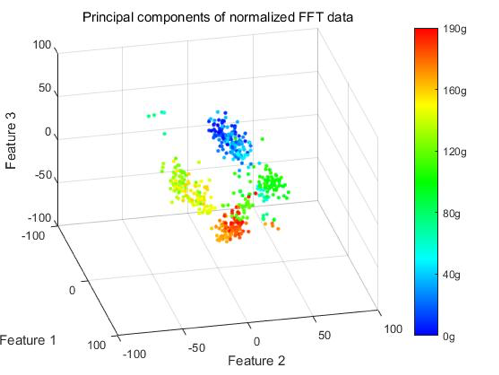}
  \caption{Principal Component of Normalized FFT Data }
\end{figure} 

\begin{figure}[h!]
  \includegraphics[width=0.4\textwidth]{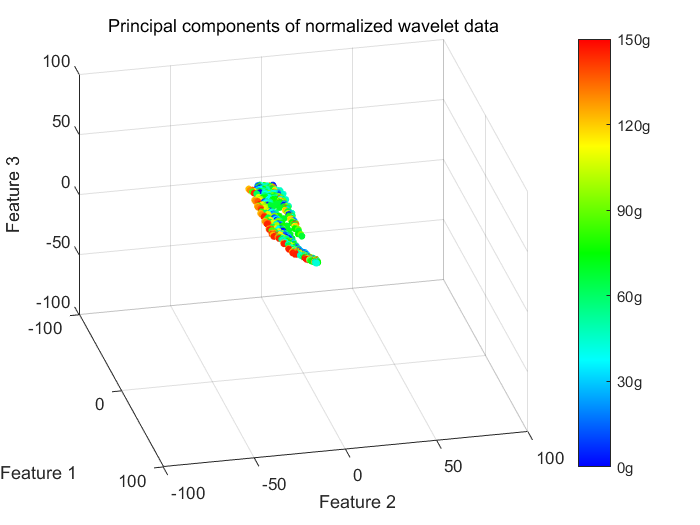}
  \caption{Principal Component of Normalized wavelet Data }
\end{figure} 

\flushleft{The colors from blue to red indicate the increasing mass from 0g to 190g as demonstrated on the color bars. The figure shows how the features change as the severity of damage (added mass) increases. The application of the feature plots is to deduce the condition of the bridge from the position of the features and to quantify the damage.}

\subsection{Support Vector Regression Model Accuracy}

\flushleft{Mean Squared Error (MSE) is used to quantify the confidence level of the outcome estimation for each model. The MSE of an estimator measures the average of the squares of the errors or deviations, that is, the difference between the ground truth and the estimation [1]. The expected uncertainty of the outcome estimation would then be the square root of the MSE. Using different Kernel functions makes assumptions on the shape of the dataset, as specific Kernel functions will fit some datasets more than others. However, by testing multiple Kernel functions and comparing the MSEs of each model, the issues of making such assumptions are avoided as the model with the lowest MSE would have the best fitting Kernel function for the dataset.}

\flushleft{Figures 6-8 show the MSEs for each kernel function used, for each dataset. When comparing MSEs of the nine different models, it is clear that the wavelet models were by far the worst performers (as expected), while the FFT models slightly outperformed the acceleration models. The Gaussian Kernel outperformed the Polynomial and Linear Kernels, and the Polynomial models outperformed the Linear models. Overall, the FFT model using the Gaussian Kernel function had the smallest MSE, with an expected error of approximately 16 grams. This model meets the project goal of getting an expected error below 10\% of the highest mass used on the bridge (190 grams).}

\flushleft{The MSE of the models will slightly vary on each run since the kernel scale is chosen heuristically, but the differences are somewhat negligible as the square root of the MSE (the expected uncertainty) usually varies by less than 1.}

\begin{figure}[h!]
  \includegraphics[width=0.3\textwidth]{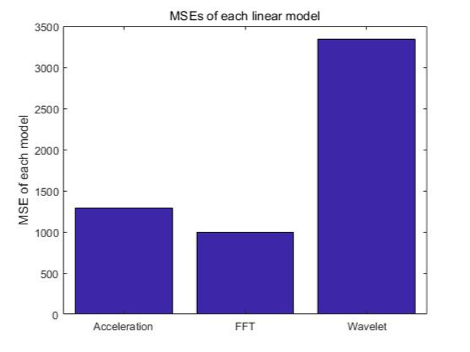}
  \caption{MSEs of Each Linear Model }
\end{figure} 
\begin{figure}[h!]
  \includegraphics[width=0.3\textwidth]{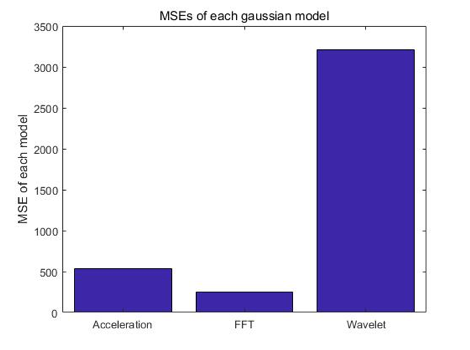}
  \caption{MSEs of Each Gaussian Model }
\end{figure} 
\begin{figure}[h!]
  \includegraphics[width=0.3\textwidth]{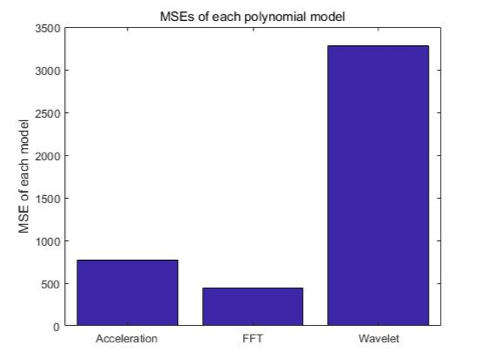}
  \caption{MSEs of Each Polynomial Model }
\end{figure} 

\subsection{Gaussian Process Regression Model Accuracy}
\flushleft{Similarly to SVR, GPR allowed the use of several Kernel functions, and a total of ten functions were considered in this project. Figure 9 shows the MSEs for the acceleration GPR models, while Figure 10 shows the MSEs for the FFT GPR models. Given how poorly the wavelet data performed in the SVR models, no GPR models were made using that data.
The MSEs of the GPR models were comparable with the MSEs of the SVR models, with the best GPR models outperforming the best SVR model. When using acceleration data, the best GPR model used the ARD Rational Quadratic Kernel function. It had an MSE of 169, which represents an expected error of 13 grams. Only two of the ten models did not meet the project's expected error goal.}

\begin{figure}[h!]
  \includegraphics[width=0.5\textwidth]{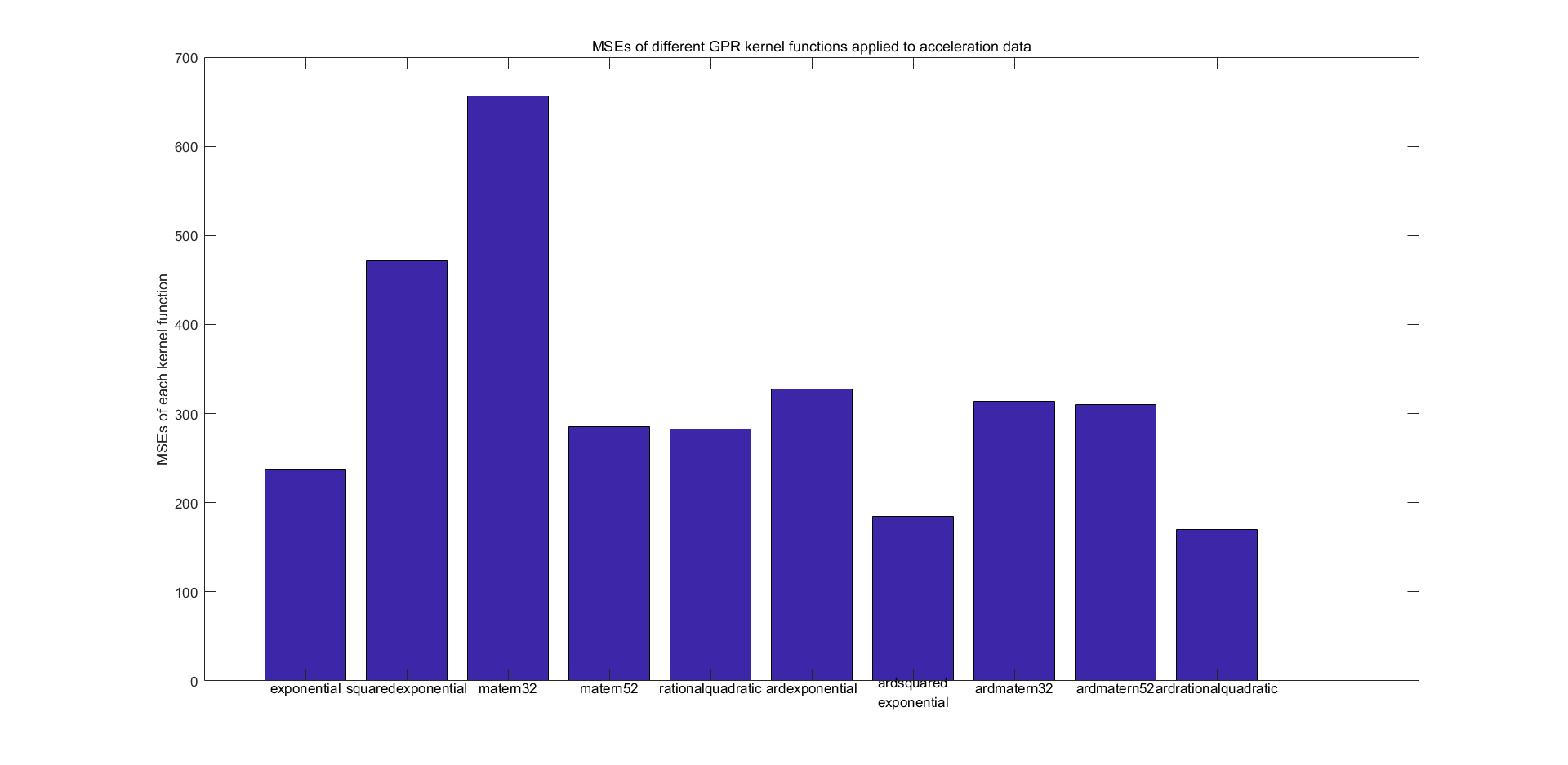}
  \caption{MSEs of Different GPR Kernel Functions applied to acceleration data}
\end{figure} 

\flushleft{When using FFT data, the best GPR model used the Mat\'ern 5/2 Kernel function. It had an MSE of 184, which represents an expected error of approximately 13.5 grams. All of the FFT models met the project's goal, however the best performing model was not as good as the best acceleration model. It is interesting to note that no Kernel function was a top three performer in both the FFT and acceleration models.}

\begin{figure}[h!]
  \includegraphics[width=0.5\textwidth]{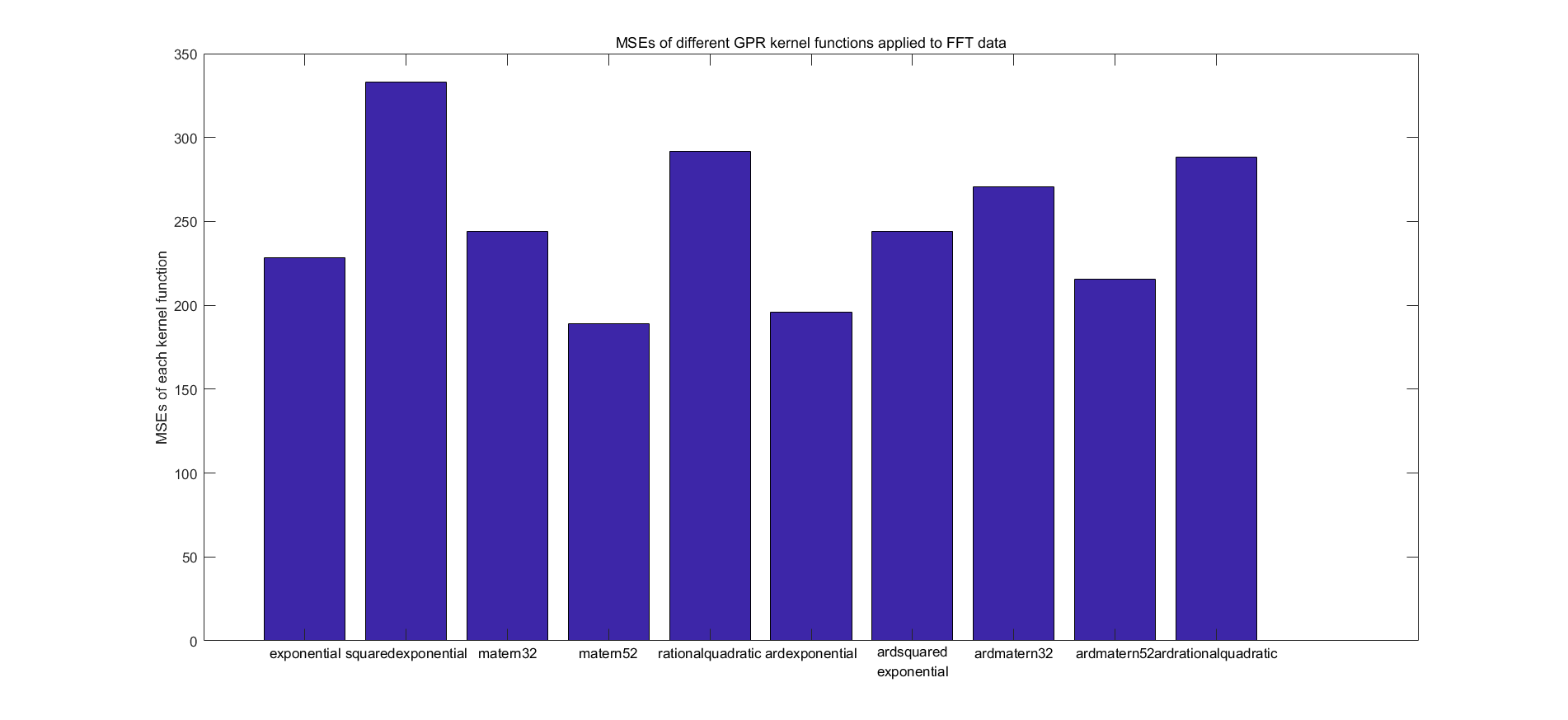}
  \caption{MSEs of Different GPR Kernel Functions applied to FFT data}
\end{figure} 

\subsection{Combined Acceleration and FFT Model Accuracy}
\flushleft{
In an attempt to include more information when training the models, the acceleration principal components were combined with the FFT principal components so that six principal components were used in the machine learning process. The wavelet principal components were not included since the dataset has a different dimension, and since it did not perform well on its own it likely would have confounded the machine learning and made the results worse. Figure 11 shows the MSEs of the SVR models using this combined data, while Figure 12 shows the MSEs of the GPR models.

All of the different Kernel functions for SVR performed better using the combined data as compared to using only the FFT or acceleration principal components. The best Kernel function is still the Gaussian function, and the expected error for that function is approximately 14 grams. 

For the GPR models, the overall performance of the models improved however there were two cases in which a Kernel function's performance was higher for acceleration or FFT data instead of the combined dataset. The best GPR model for the combined dataset was ARD Mat\'ern 3/2, which had an expected error of approximately 10 grams. This model was by far the best performing model, and there were two other GPR models with expected errors of approximately 11.5 grams.
}
%recommend that when this project is conducted on the field, a similar process of picking the best regression model and kernel function is conducted.

\begin{figure}[h!]
  \includegraphics[width=0.4\textwidth]{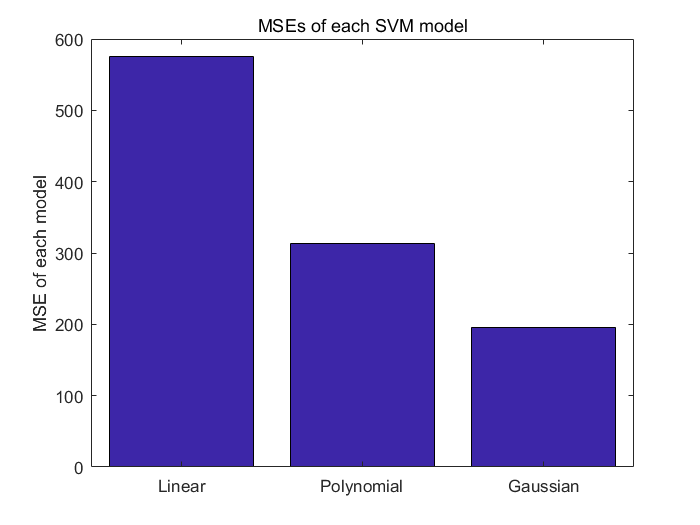}
  \caption{MSEs of combined acceleration and FFT models}
\end{figure} 

\begin{figure}[h!]
  \includegraphics[width=0.5\textwidth]{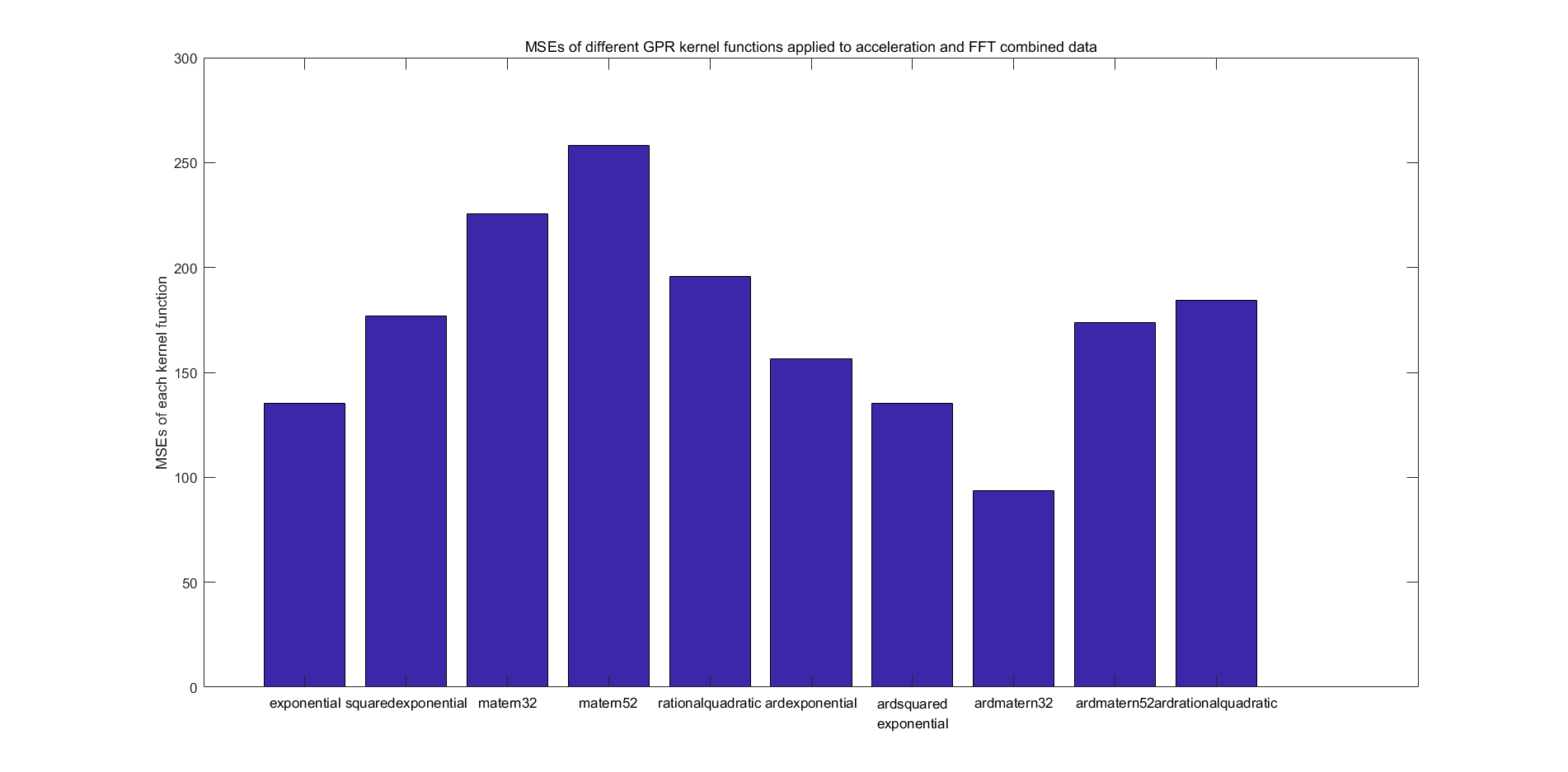}
  \caption{MSEs of Different GPR Kernel Functions applied to acceleration and FFT combined data}
\end{figure} 

\section{Limitations}
%Discuss the important challenges and issues associated with implementing all the aspects of the project, e.g., economic pressures,liability, effects of false indication of damage, hardware/software integration, verifying damage detection capability, regulatory agency oversite, etc. 
\flushleft{
NLPCA would have consistent results if a constant random seed was chosen. Additionally, SVR and GPR might perform better if optimized model hyper-parameters were found. Additionally, if there was a way in which these machine learning methods could prioritize the higher variance features, they would be better trained. Currently, all three principal components are weighed the same, however the first principal component is one with the most explained variance. It is also possible that by removing outlying points, the models' performance would have increased.
}
\section{Discussion}

\flushleft{
Multiple models were created using different features as well as different Kernel functions. Out of the 42 models created, only nine of them did not meet the goal of having an expected error below 19 grams. This is also including the three models created using the wavelet principal components, which were expected to perform poorly given the unclear clustering in the NLPCA.

Overall, the GPR models slightly outperformed the SVR models, and this is especially apparent when comparing the combined dataset GPR models with the combined dataset SVR models. Only two of the ten GPR models underperformed when compared to the best SVR model. On top of which, the best GPR model using only one set of principal components was slightly better than the best SVR model using both sets of principal components. Perhaps if more Kernel functions were tested in SVR then a better fitting one could be found.

As a whole, combining the principal components of the FFT and acceleration data proved to be better than analyzing them separately. The overall best model (GPR using the combined datasets and the ARD Mat\'ern 3/2 Kernel) had an expected error of approximately 10 grams, which is roughly half of the project's original goal.

The performance of each GPR Kernel seemed to vary for each dataset, whereas for SVR it was very clear which Kernels best fit the datasets. When applying this methodology to the field, it is recommended to approach Kernel selection similarly to how it was done in this paper to ensure the best model is chosen.
\\}

% Appendix
%\appendix
%\section{Matlab Code}
\begin{acks}
We would like to acknowledge Jingxiao Liu, who was a great help throughout our project's development.
\end{acks}

% Bibliography
\bibliographystyle{plain}
\bibliography{Final_report}
%\section{References}
\flushleft{
\textbf{[1]} Yang, Y.b., and C.w. Lin. "Vehicle-Bridge Interaction Dynamics and Potential Applications." Journal of Sound and Vibration, vol. 284, no. 1-2, 2005, pp. 205-226., doi:10.1016/j.jsv.2004.06.032. 
\flushleft{\textbf{[2]} Charles R. Farrar and Keith Worden. "An Introduction to Structural Health Monitoring". Philosophical Transactions: Mathematical, Physical and Engineering Sciences, Vol.365, No. 1851, Structural Health Monitoring (Feb. 15, 2007), pp. 303-315
\flushleft{\textbf{[3]} Mistras. (2018).MODEL 5102 High output shear design accelerometer http://www.mistrasgroup.com/products/solutions/
vibration/accelerometers/5102.aspx. Retrieved May 1st,2018
\flushleft{\textbf{[4]} Bitetto, Alessandro et al. "A Nonlinear Principal Component Analysis to Study Archeometric Data." Journal of Chemometrics 30.7 (2016): 405.
\flushleft{\textbf{[5]} Ethem Alpaydin. 2014. Introduction to Machine Learning. The MIT Press.
\flushleft{\textbf{[6]} Lehmann, E. L.; Casella, George (1998). Theory of Point Estimation (2nd ed.). New York: Springer. ISBN 0-387-98502-6. MR 1639875
\flushleft{\textbf{[7]} Machine Learning with MATLAB-Google AdWords - Confirmation. MATLAB \& Simulink, www.mathworks.com/campaigns/products/ppc/google/machine-learning-with-matlab-conf.html?elqsid=1523566036978 \& potential-use=Student.
\flushleft{\textbf{[8]} Alex J. Smola; Bernhard Scholkopf(2003). A Tutorial on Support Vector Regression. Netherlands:Statistics and Computing 14:199 - 222, 20 (https://alex.smola.org/papers/2003/SmoSch03b.pdf)
\flushleft{\textbf{[9]} Bishop, Christopher M. Pattern Recognition and Machine Learning. Springer, 2006.
\flushleft{\textbf{[10]} D.Basak, S. Pal and D.C. Patranabis, "Support Vector Regression," 12 July 2007. [Online]. Available (https://pdfs.semanticscholar.org/
c5a9/67eaded74a9fc414de4ad5120b0b66acd2c3.pd f)
\flushleft{\textbf{[11]} R.Dingledine, R.Lethinhttp, "Use of Support Vector Machines for Supplier Performance Modeling," [Online]. Available: (//web.mit.edu/arma/Public/svm.pdf)
\flushleft{\textbf{[12]} Linting, M.,Meulman, J.J., Groenen, P.J.F., \& Van der Kooij, A.J. (2007). Nonlinear principal componentsanalysis: Introduction and application. Psychological Methods. In press.
\flushleft{\textbf{[13]} Ebden, M. "Generalized Gaussian Process Regression for Non-Gaussian Functional Data." Gaussian Process Regression Analysis for Functional Data, 2011, pp. 119-138., doi:10.1201/b11038-8.
\flushleft{\textbf{[14]} Zhou, Zhi, and Jinping Ou. "Development of FBG Sensors for Structural Health Monitoring in Civil Infrastructures." Sensing Issues in Civil Structural Health Monitoring, pp. 197-207., doi:10.1007/1-4020-3661-2-20.
\flushleft{\textbf{[15]} Agdas, Duzgun, et al. "Comparison of Visual Inspection and Structural-Health Monitoring As Bridge Condition Assessment Methods." Journal of Performance of Constructed Facilities, vol. 30, no. 3, 2016, p. 04015049., doi:10.1061/(asce)cf.1943-5509.0000802.
\flushleft{\textbf{[16]} Lederman, G, et al. "Damage Quantification and Localization Algorithms for Indirect SHM of Bridges." Bridge Maintenance, Safety, Management and Life Extension, 2014, pp. 640-647., doi:10.1201/b17063-93}

\end{document}